\documentclass{article}

     \usepackage[final]{neurips_rl2019}

\usepackage[utf8]{inputenc} 
\usepackage[T1]{fontenc}    
\usepackage{hyperref}       
\usepackage{url}            
\usepackage{booktabs}       
\usepackage{amsfonts}       
\usepackage{nicefrac}       
\usepackage{microtype}      
\usepackage{amsfonts}
\usepackage{graphicx}
\usepackage{algorithm}
\usepackage{algorithmic}
\usepackage{bbm}
\usepackage{bm}
\usepackage{amsmath}
\usepackage{wrapfig}
\usepackage{subcaption}
\usepackage{caption}
\usepackage{amssymb}
\usepackage[dvipsnames]{xcolor}

\title{Mutual Information Maximization for Robust Plannable Representations}

\author{Yiming Ding, Ignasi Clavera, Pieter Abbeel \\
University of California, Berkeley \\
\texttt{\{dingyiming0427, iclavera, pabbeel\}@berkeley.edu}
}

\begin{document}

\maketitle

\begin{abstract}
Extending the capabilities of robotics to real-world complex, unstructured environments requires the need of developing better perception systems while maintaining low sample complexity. When dealing with high-dimensional state spaces, current methods are either model-free or model-based based on reconstruction objectives. The sample inefficiency of the former constitutes a major barrier for applying them to the real-world. The later, while they present low sample complexity, they learn latent spaces that need to reconstruct every single detail of the scene. In real environments, the task typically just represents a small fraction of the scene. Reconstruction objectives suffer in such scenarios as they capture all the unnecessary components. In this work, we present MIRO, an information theoretic representational learning algorithm for model-based reinforcement learning. We design a latent space that maximizes the mutual information with the future information while being able to capture all the information needed for planning. We show that our approach is more robust than reconstruction objectives in the presence of distractors and cluttered scenes\footnote{Accepted at NeurIPS 2019 Workshop on Robot Learning:
Control and Interaction in the Real World}.
\end{abstract}
\vspace{-0.1 in}
\section{Introduction}
\vspace{-0.1 in}




A fundamental challenge in applying reinforcement learning (RL) to real robotics is the need to define a suitable state space representation. Designing perception systems manually makes it difficult to apply the same RL algorithm to a wide range of manipulation tasks in complex, unstructured environments. One could, in principle, apply model-free reinforcement algorithms from raw, low-level observations; however, these tend to be slow, sensitive to hyperparameters, and sample inefficient~\cite{Lee2019}. Model-based reinforcement learning, where a predictive model of the environment is first learned and then a controller is derived from it, offers the potential to develop sample efficient algorithms even from raw, low-level observations. However, Model-based RL from high-dimensional observations poses a challenging problem: How do we learn a good latent space for planning?

Recent work in learning latent spaces for model-based RL can be categorized in three main classes: 1) hand-crafted latent spaces ~\cite{Finn2016}, 2) video prediction models~\cite{Jayaraman2018}, and 3) latent space learning with reconstruction objectives~\cite{hafner2018learning, Lee2019, Watter2015, Wahlstrom2015, Zhang2018, Ha2018}. While hand-crafted latent spaces provide a good inductive bias, such as forcing the latent to be feature points~\cite{Finn2016}, they are too rigid for unstructured tasks and cannot incorporate semantic knowledge of the environment. Video prediction models and latent space with reconstruction objectives share the commonality that the latent space is learned using loss on the raw pixel observations. As a result, the latent space model needs to incorporate all information to reconstruct every detail in the observations, which is redundant as the task is usually represented by a small fraction of the scene in real world environments.

Instead, this work tackles the problem of representation learning from an information theoretic point of view: learning a latent space that maximizes the mutual information (MI) between the latent and future observations. The learned latent space encodes the underlying shared information between the different future observations, discarding the low-level information, and local noise. When predicting further into the future, the amount of information becomes much lower and the models needs to infer the global structure. The MI is optimized using energy-based models and a discriminative procedure~\cite{Oord2018, Henaff2019}. Energy-based models do not pose any assumption on the distribution between the latent and the image, and the discriminative procedure results in a robust latent space.

The main contribution of our work is a representational learning approach, MIRO (\textbf{M}utual \textbf{I}nformation \textbf{Ro}bust Representation), which jointly optimizes for the latent representation and model, and results in a latent space that is robust against disturbances, noisy observations, and can achieve performance comparable to state-of-the-art model-based algorithms. Our experimental evaluation illustrates the strengths of our framework on four standard DeepMind Control Suite~\cite{tassa2018deepmind} environments. 
\vspace{-0.1 in}
\section{Background}
\vspace{-0.1 in}
This work tackles the problem of learning a latent space that is suitable for planning from high-dimensional observations in POMDPs. We maximize the mutual information (MI) of the latent variables and the future observations, and learn a predictive model of the environment.

\textbf{Partially Observable Markov Decision Process. }
A discrete-time finite partially observable Markov decision process (POMDP) $\mathcal{M}$ is defined by the tuple $(\mathcal{S}, \mathcal{A}, T, R, \mathcal{O}, O, \gamma, \rho_0, H)$. 
Here, $\mathcal{S}$ is the set of states, $\mathcal{A}$ the action space, $T(s_{t+1}| s_t, a_t) = p(s_{t+1}|s_t, a_t)$ the transition distribution, $R(s_t) = p(r_t|s_t)$ is the probability of obtaining the reward $r_t$ at the state $s_t$, $\mathcal{O}$ is the observation space, $O(o_t|s_t) = p(o_t|s_t)$,  $\gamma$ the discount factor, $\rho_0: \mathcal{S} \to \mathbb{R}_+$ represents the initial state distribution, and $H$ is the horizon of the process. We define the return as the sum of rewards $r_t$ along a trajectory $\tau := (s_{0}, a_{0}, ..., s_{H-1}, a_{H-1}, s_{H})$. The goal of reinforcement learning is to find a controller $\pi: \mathcal{S} \times \mathcal{A} \rightarrow \mathbb{R}^+$ that maximizes the expected return, i.e.:
    $\max_{\pi} J(\pi) = \mathbb{E}_{\begin{subarray}{l}{
a_t \sim \pi}\\
s_{t} \sim p
\end{subarray}}
[\sum_{t=1}^{H}\gamma^{t} r_t]$.


\textbf{Mutual Information. }
The mutual information between two random variable $X$ and $Y$, denoted by $I(X;Y)$ is a reparametrization-invariant measure of dependency. Specifically, it characterizes the Kullback-Leibler divergence between the joint distribution $(X, Y)$ and the product of the marginals $X$ and $Y$:
$I(X; Y) = \mathbb{E}_{p(x, y)} \left[ \log \frac{p(x|y)}{p(x)} \right]  = \mathbb{E}_{p(x, y)} \left[ \log \frac{p(x, y)}{p(x) p(y)} \right] = D_{\text{KL}}((X,Y)\| X \otimes Y) $.

Estimating and optimizing the mutual information objective poses a challenging problem. In this work, we use a multi-sample
unnormalized lower bound based on noise contrastive estimation, $I_{\text{NCE}}$~\cite{Henaff2019}. This estimator is a lower bound on the mutual information.


\vspace{-0.1 in}
\section{Method}
\vspace{-0.1 in}
Enabling complex real robotics tasks requires extending current model-based methods to low-level high-dimensional observations. However, in order to do so, we need to specify which space we should plan with. Real-world environments are unstructured, cluttered, and present distractors. Our approach, MIRO, is able to learn latent representations that capture the relevant aspects of the dynamics and the task by framing the representational learning problem in information theoretic terms: maximizing the mutual information between the latent space and the future observations. This objective advocates for representing just the aspects that matter in the dynamics, removing the burden of reconstructing the entire pixel observation.

\subsection{Learning Latent Spaces for Control}
Since the learned latent space will be used for planning, it has to be able to accurately predict the rewards and incorporate new observations when available. To this end, we decompose the latent space learning in a single optimization procedure  composed of four functions:

\textbf{Encoder. } The encoder is a function that maps high-dimensional observation to a lower dimensional manifold, $e_{\phi}: \mathcal{O} \to \mathcal{Z} \subseteq \mathbb{R}^m$. We represent the encoded observation $e_{\phi}(o_t)$ with the notation $z_t \in \mathcal{Z}$. Contrary to prior work in learning latent spaces for planning, we do not make any assumption on the underlying distribution of $z_t$.

\textbf{Dynamics model. } The dynamics model gives the transition function on the latent space, $f_{\theta}: \mathcal{S} \times \mathcal{A} \to \mathcal{S}$; it generates the next latent state $\hat{s}_{t+1}$ given the the current state $s_t$ and action $a_t$. The dynamics model is probabilistic following a Gaussian distribution, i.e., $s_{t+1} \sim \mathcal{N}(\mu_t, \sigma^2_t)$. 

\textbf{Filtering function. } This function $g_{\zeta}: \mathcal{Z} \times \mathcal{S} \to \mathcal{S}$ filters the belief of the latent state $\hat{s}_t$ with the current encoded observation $z_t$ to obtain the filtered latent variable $s_t$.

\textbf{Reward predictor. } This function is a mapping between latent states and rewards, $r_{\psi}: \mathcal{S} \to \mathcal{A}$. We assume that the rewards follow a Gaussian distribution with unit variance.

The parameters of these functions are learned in a single optimization objective; specifcally, we optimize the following constrain optimization problem:
\begin{align*}
     \max_{\bf{\theta, \phi, \psi, \zeta}} \hspace{0.15 in}  I(s_t; o_{t+h} | a_t, \dots, a_{t+h-1}) \hspace{1.2 in} \\
\text{s.t.:} \quad \quad  D_{\text{KL}}(g_{\zeta}(z_{t+1}, \hat{s}_{t+1}) \| s_{t+1}) = 0 \quad \quad z_t = e_{\phi}(o_t) \\
     D_{\text{KL}}(r_t \| r_{\psi}(s_t, a_t)) = 0 \hspace{.35 in} \hat{s}_{t+1} \sim f_{\theta}(s_t, a_t) \\
     \vspace{-0.2 in}
\end{align*}
The mutual information term preserves in latent space only the information relevant for the dynamics, while the constraints prevents the latent from degenerating on the terms that matter to predict the rewards and next latents. However, in practice, this objective is intractable: we cannot evaluate the exact mutual information term and we have a set of non-linear constraints. To optimize it, we formulate the Lagrangian of the objective and replace the mutual information objective with the noise contrastive estimator lower bound $I_{\text{NCE}}$:
\begin{align}
    \max_{\bf{\theta, \phi, \psi, \zeta}} I_{\text{NCE}}(s_t; o_{t+h} | a_t, \dots, a_{t+h-1}) - \lambda_1 D_{\text{KL}}(g_{\zeta}(z_t, \hat{s}_{t+1}) \| \hat{s}_{t+1}) - \lambda_2 D_{\text{KL}}(r_t \| r_{\psi}(s_t, a_t))
\label{eq:loss}
\end{align}
We use the reparametrization trick~\cite{kingma2013auto} for $\hat{s_{t+1}} = \hat{\mu}_{t+1} + \xi \hat{\omega}_{t+1}$ where $\xi \sim \mathcal{N}(0, 1)$ and $z_t = e_{\phi}(o_t)$. In our case, the noise contrastive estimator term results in:
\begin{align}
    I_{\text{NCE}}(s_t; o_{t+h} | a_t, \dots, a_{t+h-1}) = \mathbb{E}\left[\log \left(\frac{\text{exp}(\hat{s}_{t+h}^\top W_h z_{t+h})}{
    \text{exp}(\hat{s}_{t+h}^\top W_h z_{t+h}) + \sum_{j=1}^{K-1} \text{exp}(\hat{s}_{t+h}^\top W_h z_{j})} \right) \right]
\end{align}
Here, $\hat{s}_{t+h}$ represents the open loop prediction from $s_t$ and the sequences of actions $a_{t}, \dots, a_{t+h}$.

\subsection{Latent Space Model-Based Reinforcement Learning}
We present the overall method for obtaining a model-based controller when learning the latent space using MIRO. 

\textbf{Data Collection. } As is typical of model-based methods, we alternate between model learning and data collection using the latest controller. This allows the model to just learn the parts of the state space that the agent visits, removing the burden of modelling the entire space, and overcomes the insufficient coverage of the initial data distribution.

\textbf{Model Learning. } We learn the encoder, model, filtering function, and reward predictor altogether using equation~\ref{eq:loss} and all the data collected so far. 

\textbf{Planning. } As a controller, we use model-predictive control (MPC) with cross-entropy method (CEM) \cite{botev2013cross}. The CEM component selects the action that maximizes the expected return under the learned models. Specifically, it is a population based procedure that iteratively refits a Gaussian distribution on the best sequences of actions. The MPC component results in a more robust planning by doing CEM at each step.

\vspace{-0.1 in}
\section{Results}
\vspace{-0.1 in}
In this section, we empirically corroborate the claims in the previous sections. Specifically, the experiments are designed to address the following questions: 
1) Is our approach able to maintain its performance in front of distractors in the scene? 
2) How does our method compare with state-of-the-art reconstruction objectives? 

To answer the posed questions, we evaluate our framework, in four continuous control benchmark tasks MuJoCo simulator: cartpole balance, reacher, finger spin and half cheetah ~\cite{2012mujoco, tassa2018deepmind}. We choose PlaNet \cite{hafner2018learning} as the state-of-the-art reconstruction objective baseline. 

\begin{figure}[t]
    \centering
    \includegraphics[width=\linewidth]{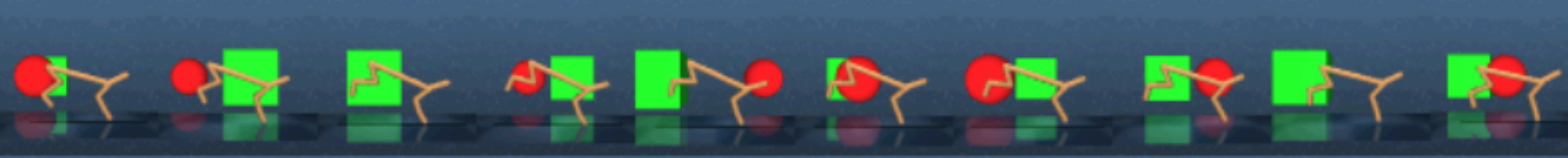}
    \caption{ \small An example trajectory of Cheetah environment with distractors. The distractor objects (red sphere and green cube) are placed at a random position at each time step.}
    \label{fig:noisy_cheetah}
    \vspace{-0.2 in}
\end{figure}

\begin{figure}[h]
    \centering
    \begin{subfigure}[H]{0.4\linewidth}\centering
    \includegraphics[width=\linewidth]{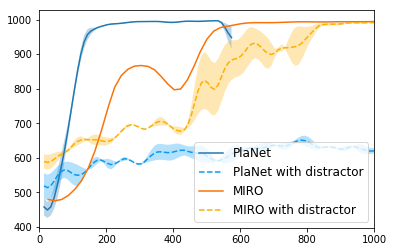}
    \caption{Cartpole Balance}
    \end{subfigure}
    \begin{subfigure}[H]{0.4\linewidth}\centering
    \includegraphics[width=\linewidth]{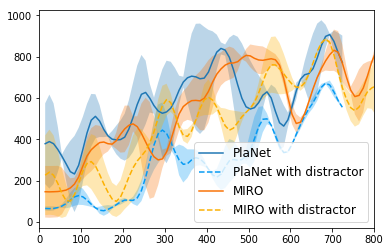}
    \caption{Reacher}
    \end{subfigure}
    \\
    \begin{subfigure}[H]{0.4\linewidth}\centering
    \includegraphics[width=\linewidth]{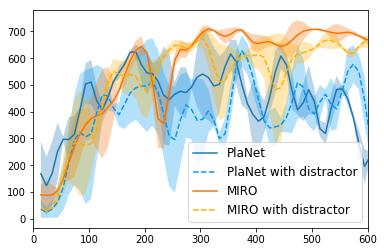}
    \caption{Finger}
    \end{subfigure}
    \begin{subfigure}[H]{0.4\linewidth}\centering
    \includegraphics[width=\linewidth]{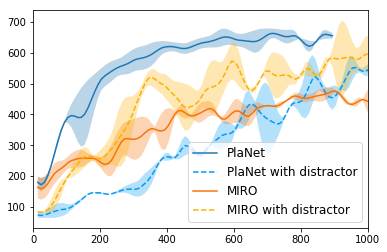}
    \caption{Half Cheetah}
    \end{subfigure}
    \caption{\small Learning curves of MIRO and PlaNet on environments with and without distractors. All curves represent mean and the shaded area represents one standard deviation among 3 seeds.}
    \label{fig:learning_curves}
    \vspace{-0.1 in}
\end{figure}

\textbf{Robustness Against Distractors. } To test the robustness of MIRO and PlaNet in visually noisy environments, we add distractors to each of the four above environments in the background as shown in Fig. \ref{fig:noisy_cheetah}. As shown in Fig. \ref{fig:learning_curves}, we observe that in all four environments, the performance of MIRO is not undermined by the presence of distractors. Interestingly, in the Cheetah environment, the performance even improves with visual noise in the background. A possible explanation is that the presence of the distractors forces the embedding to focus more on information relevant to the task and thus makes the embedding more suitable for planning.  In comparison, the performance of PlaNet struggles in face of distractors. The agent barely learns in the Cartpole Balance environment, and in the Cheetah environment it suffers from a slower takeoff.

\textbf{Comparison Against Reconstruction Objectices} When no distractors ae present int the secene, we see that MIRO is able to achieve the same or superior asymptotic performance than its reconstruction-based counterpart method. Its learning speed its environment dependent. However, given the the performance boost brought by distractors, further regularization of the learned latent space presents it self as a promising direction towards performance improvement.



\section{Conclusion}
In this paper, we present MIRO, a mutual information based representation learning approach for model-based RL. Compared to state-of-the-art reconstruction-based methods, MIRO is more robust in noisy visual background in the environment, which is prevalent in real world applications. The initial results shed light on the importance of information theoretical representation learning and open an enticing research direction. Our next steps include optimizing the sampling of negative samples in NCE objective to encourage the latent space to disregard task-irrelevant information.

\section*{Acknowledgments}
This work was supported in part by Berkeley Deep Drive (BDD) and ONR PECASE N000141612723.

\clearpage
\bibliography{ref,mbcpc}

\begin{thebibliography}{10}

\bibitem{botev2013cross}
Z.~I. Botev, D.~P. Kroese, R.~Y. Rubinstein, and P.~L’Ecuyer.
\newblock The cross-entropy method for optimization.
\newblock In {\em Handbook of statistics}, volume~31, pages 35--59. Elsevier,
  2013.

\bibitem{Finn2016}
C.~Finn, X.~Y. Tan, Y.~Duan, T.~Darrell, S.~Levine, and P.~Abbeel.
\newblock {Learning Visual Feature Spaces for Robotic Manipulation with Deep
  Spatial Autoencoders}.
\newblock {\em Proceedings - IEEE International Conference on Robotics and
  Automation}, 2016-June:512--519, 2016.

\bibitem{Ha2018}
D.~Ha and J.~Schmidhuber.
\newblock World models.
\newblock {\em CoRR}, abs/1803.10122, 2018.

\bibitem{hafner2018learning}
D.~Hafner, T.~Lillicrap, I.~Fischer, R.~Villegas, D.~Ha, H.~Lee, and
  J.~Davidson.
\newblock Learning latent dynamics for planning from pixels.
\newblock {\em arXiv preprint arXiv:1811.04551}, 2018.

\bibitem{Henaff2019}
O.~J. H{\'{e}}naff, A.~Srinivas, J.~D. Fauw, A.~Razavi, C.~Doersch, S.~M.~A.
  Eslami, and A.~van~den Oord.
\newblock Data-efficient image recognition with contrastive predictive coding.
\newblock {\em CoRR}, abs/1905.09272, 2019.

\bibitem{Jayaraman2018}
D.~Jayaraman, F.~Ebert, A.~A. Efros, and S.~Levine.
\newblock Time-agnostic prediction: Predicting predictable video frames.
\newblock {\em CoRR}, abs/1808.07784, 2018.

\bibitem{kingma2013auto}
D.~P. Kingma and M.~Welling.
\newblock Auto-encoding variational bayes.
\newblock {\em arXiv preprint arXiv:1312.6114}, 2013.

\bibitem{Lee2019}
A.~X. Lee, A.~Nagabandi, P.~Abbeel, and S.~Levine.
\newblock Stochastic latent actor-critic: Deep reinforcement learning with a
  latent variable model.
\newblock {\em CoRR}, abs/1907.00953, 2019.

\bibitem{tassa2018deepmind}
Y.~Tassa, Y.~Doron, A.~Muldal, T.~Erez, Y.~Li, D.~d.~L. Casas, D.~Budden,
  A.~Abdolmaleki, J.~Merel, A.~Lefrancq, et~al.
\newblock Deepmind control suite.
\newblock {\em arXiv preprint arXiv:1801.00690}, 2018.

\bibitem{2012mujoco}
E.~Todorov, T.~Erez, and Y.~Tassa.
\newblock Mujoco: A physics engine for model-based control.
\newblock In {\em Intelligent Robots and Systems (IROS), 2012 IEEE/RSJ
  International Conference on}, pages 5026--5033. IEEE, 2012.

\bibitem{Oord2018}
A.~van~den Oord, Y.~Li, and O.~Vinyals.
\newblock {Representation Learning with Contrastive Predictive Coding}.
\newblock {\em CoRR}, jul 2018.

\bibitem{Wahlstrom2015}
N.~Wahlström, T.~B. Schön, and M.~P. Deisenroth.
\newblock From pixels to torques: Policy learning with deep dynamical models,
  2015.

\bibitem{Watter2015}
M.~Watter, J.~T. Springenberg, J.~Boedecker, and M.~A. Riedmiller.
\newblock Embed to control: {A} locally linear latent dynamics model for
  control from raw images.
\newblock {\em CoRR}, abs/1506.07365, 2015.

\bibitem{Zhang2018}
M.~Zhang, S.~Vikram, L.~Smith, P.~Abbeel, M.~J. Johnson, and S.~Levine.
\newblock {SOLAR: Deep Structured Representations for Model-Based Reinforcement
  Learning}.
\newblock {\em CoRR}, aug 2018.

\end{thebibliography}
\bibliographystyle{abbrv}

\end{document}